\documentclass[journal]{IEEEtran}
\usepackage{amsmath,amsfonts}
\usepackage{algorithmicx}
\usepackage{algpseudocode}
\usepackage{algorithm}
\usepackage{array}
\usepackage[caption=false,font=normalsize,labelfont=sf,textfont=sf]{subfig}
\usepackage{textcomp}
\usepackage{stfloats}
\usepackage{url}
\usepackage{verbatim}
\usepackage{graphicx}
\usepackage{cite}
\usepackage{times}
\usepackage{epsfig}
\usepackage{graphicx}
\usepackage{amssymb}
\usepackage{fontawesome}
\usepackage{multirow}
\usepackage{tabularx}
\usepackage{color}
\usepackage{fancyhdr}
\usepackage{mathtools}
\usepackage{amsthm}
\usepackage[table]{xcolor}
\usepackage{pifont}
\usepackage{booktabs}
\newcommand{\cmark}{\ding{51}}
\newcommand{\xmark}{\ding{55}}

\definecolor{myblue}{RGB}{0,92,175}

\begin{document}

\title{Causal-AgentIR: Self-Evolving Causal Memory for Adaptive Image Restoration Agents}

\author{Hu Gao, Yulong Chen and Lizhuang Ma$^{\dag}$, Senior Member, IEEE
\thanks{Hu Gao and  Lizhuang Ma are  with the Department of Computer Science, Shanghai Jiao Tong University, Shanghai 200240, China (e-mail: gao\_h@sjtu.edu.cn, lzma@sjtu.edu.cn).

Yulong Chen is with the Department of Architecture and Design, Harbin Institute of Technology, Heilongjiang, China (e-mail: llong\_c@hit.edu.cn)
}}



\maketitle

\begin{abstract}
Image restoration agents have recently emerged as a flexible paradigm for handling diverse and unpredictable degradations in real-world scenarios. Existing agents typically formulate restoration as a tool-using process, where the agent perceives degradations, searches candidate tools, executes restoration operations, and revises the plan through reflection or rollback. However, their knowledge is often stored as static tool descriptions, manually defined degradation priors, or unstructured textual summaries, which limits the accumulation, verification, revision, and forgetting of restoration knowledge over long-term experience. In this paper, we propose Causal-AgentIR, a hierarchical multi-agent framework with self-evolving causal memory for collective image restoration intelligence. Instead of representing restoration experience as isolated textual records, Causal-AgentIR organizes degradation patterns, image regions, restoration tools, actions, quality changes, and user preferences into a structured causal memory graph. This graph supports graph-based retrieval and multi-hop causal reasoning, enabling agents to infer how specific restoration operations or tool sequences affect restoration quality under different degradation conditions. The framework further organizes multiple agents into a collaborative system, including planning, degradation analysis, tool expertise, causal memory reasoning, outcome critique, and memory curation. Through this design, restoration experience can be added, updated, merged, reinforced, ignored, or discarded according to observed quality changes and feedback, allowing the agent to maintain reliable and transferable restoration knowledge. 
 Extensive experiments demonstrate the effectiveness of the proposed framework.

\end{abstract}


\section{Introduction}
Image restoration aims to recover high-quality images from degraded observations and remains a fundamental problem in low-level vision. Over the past decades, considerable progress has been made in task-specific restoration, including image denoising, deblurring, deraining, dehazing and desnowing, and related tasks~\cite{denoise10908805,ALGgao2024learning,efderainguo2025efficientderain+,PGH2Netisu2025prior,LSSRgao2024learning}. Driven by deep convolutional networks, Transformers, diffusion models, and state space models, task-specific methods have achieved strong performance on well-defined benchmarks. However, image degradations in real-world scenarios are often diverse, unpredictable, and spatially non-uniform. Such variability makes real-world restoration substantially more challenging than conventional task-specific settings with predefined degradation types.

To improve adaptability across degradation scenarios, recent studies have explored task-aligned and all-in-one image restoration. Task-aligned methods train a general restoration network across multiple tasks~\cite{FSNet,gao2025mixed,SFNet,aclgu2025acl,starir11429607,guo2025mambairv2}. Although these methods enhance model transferability, they still commonly rely on known degradation settings during inference and may struggle with unknown or domain-specific distortions. All-in-one restoration further seeks to handle multiple degradation types within a single unified framework. Representative approaches~\cite{potlapalli2023promptir,CAPTNet10526271,baryir11417902,autodir10.1007/978-3-031-73661-2_19,VLUNetZeng_2025_CVPR} use degradation embeddings, prompt learning, instruction conditioning, dynamic routing, or vision-language priors to guide the restoration process. Despite improved task coverage, these methods still face an inherent trade-off between generality and specialization: a unified model is expected to accommodate diverse degradations, yet it may be less effective than specialized models on individual tasks. Moreover, as static models trained on predefined degradation categories, they often lack the flexibility to adapt their restoration strategies to unseen degradation combinations, emerging artifacts, or user-specific requirements.
\begin{figure}[t]
\centering
\includegraphics[width=\linewidth]{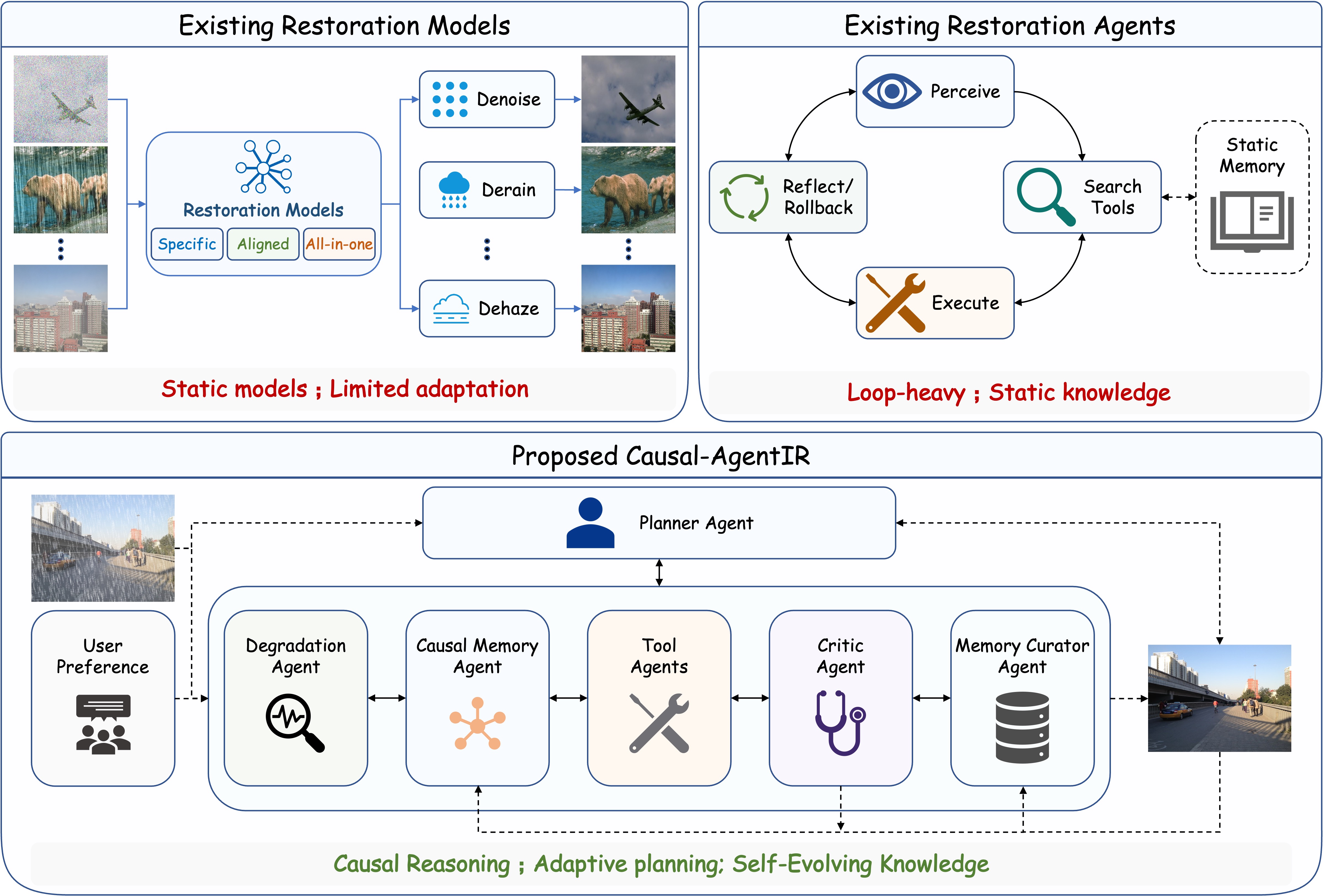}
\caption{Comparison between existing image restoration paradigms and the proposed Causal-AgentIR.}
\label{fig:exam}
\end{figure}

Motivated by the recent progress in large language models (LLMs)~\cite{LLMbrown2020language,LLMroziere2023code,LLMtouvron2023llama}  and multimodal agents~\cite{Magentxie2025large,Magentyang2025magma}, agentic image restoration has emerged as a new direction~\cite{RestoreAgent,MAIRjiang2026multi,agenticir,hybridagentli2026hybrid,wei2026iamagent,lu2026restorer1}. Rather than relying on a single monolithic model, restoration agents formulate image restoration as a decision-making process. They can analyze the input image, identify possible degradations, select appropriate restoration tools, execute operations, evaluate intermediate results, and revise the restoration plan. This paradigm allows multiple specialized and general restoration models to be coordinated within a flexible restoration workflow, while also providing a natural interface for user interaction and personalized restoration.

Despite their flexibility, existing restoration agents still rely heavily on repeated tool search, reflection, and rollback, which may introduce considerable inference overhead. Their knowledge is commonly represented by static tool descriptions, manually defined degradation priors, prompt templates, or unstructured textual summaries. Such representations can support short-term planning, but they are difficult to verify, revise, consolidate, or selectively forget as new restoration experience is accumulated. More importantly, they rarely encode context-dependent action effects, including how a restoration operation or tool sequence affects image quality under different degradation mixtures, spatial regions, and user requirements. For example, direct dehazing may amplify sensor noise, whereas mild denoising before dehazing may produce a more reliable restoration trajectory. Consequently, existing agents remain effective tool-using systems but provide limited support for long-term restoration knowledge evolution.

To address this limitation, we propose Causal-AgentIR, a hierarchical multi-agent framework with self-evolving causal memory for adaptive image restoration, as shown in Figure~\ref{fig:exam}. Instead of storing restoration experience as isolated textual records, Causal-AgentIR organizes degradations, image regions, restoration tools, actions, quality changes, computational costs, and user preferences into a structured memory graph. The graph represents empirical and context-dependent action--outcome relations obtained from executed restoration operations. During inference, relevant subgraphs are retrieved to support restoration planning, tool selection, regional processing, and execution-order determination. After each operation, restoration outcomes and feedback are evaluated to revise the corresponding memory relations. Through hierarchical collaboration among the planner, degradation agent, tool agents, causal memory agent, critic, and memory curator, restoration experience can be continuously reused, verified, consolidated, reinforced, or discarded.

Table~\ref{tab:agent_comparison} summarizes the main differences between representative restoration agents and Causal-AgentIR. Unlike existing systems that primarily focus on tool invocation, multi-agent coordination, or trajectory-level policy learning, Causal-AgentIR explicitly models structured restoration memory and its continual evolution.

\begin{table*}
\centering
\caption{Comparison between representative image restoration agents and the proposed Causal-AgentIR.}
\label{tab:agent_comparison}
\resizebox{\textwidth}{!}{
\begin{tabular}{ccccccc}
\toprule
Method & Multi-Agent & Graph Memory & Causal Reasoning & Self-Updated Memory  & Human Feedback & Tool Extensibility \\
\hline
\hline
MAIR~\cite{MAIRjiang2026multi}     & \cmark & \xmark & prior-based & \xmark & \cmark    & high \\
AgenticIR~\cite{agenticir}        & \xmark & \xmark & weak        & \xmark  & \xmark    & medium \\
HybridAgent~\cite{hybridagentli2026hybrid} & \cmark & \xmark & \xmark & \xmark & \cmark & medium \\
IAMAgent~\cite{wei2026iamagent}   & \cmark & \xmark & \xmark       & \xmark & \cmark    & high \\
Restore-R1~\cite{lu2026restorer1}  & \xmark & \xmark & \xmark      & \xmark & implicit  & medium \\
\hline
\rowcolor{gray!20} \textbf{Causal-AgentIR}            & \cmark & \cmark       & \cmark & \cmark & \cmark    & high \\
\hline
\end{tabular}
}
\end{table*}

Our contributions are summarized as follows:
\begin{itemize}
\item We propose Causal-AgentIR, which formulates agentic image restoration as a sequential decision-making process that jointly integrates restoration planning and post-execution memory evolution within a hierarchical multi-agent system.

\item  We introduce a self-evolving causal memory graph together with a learnable memory evolution mechanism. The graph encodes context-dependent action effects, execution order, quality variation, cost, and user preference, while supporting structured retrieval, multi-hop reasoning, evidence consolidation, and selective forgetting.

\item Extensive experiments across task-specific, all-in-one, real-world, and mixed-degradation benchmarks demonstrate the effectiveness and generality.
\end{itemize}

\section{Related Work}

\subsection{Image Restoration}

Image restoration aims to recover high-quality images from degraded observations and remains a fundamental problem in low-level vision. Early methods mainly relied on handcrafted priors, such as sparsity constraints, low-rank assumptions, and physical degradation models~\cite{2011Single,10558778,11342300}. Although interpretable, these methods are limited by manually designed assumptions and often generalize poorly to complex real-world degradations. Recent deep learning methods have substantially improved restoration performance by learning degradation-aware representations from large-scale data~\cite{LSSRgao2024learning,FSNet,FDTANetgao2025frequency}. ALGNet~\cite{ALGgao2024learning} performs adaptive local-global feature extraction for motion deblurring, XYScanNet~\cite{liu2024xyscannet} uses alternating scanning to capture long-range spatial dependencies, EfDeRain+~\cite{efderainguo2025efficientderain+} formulates deraining as predictive filtering, and PGH$^2$Net~\cite{PGH2Netisu2025prior} integrates bright/dark channel priors with histogram equalization for hierarchical dehazing. Generative restoration methods, such as UPID-EDM~\cite{upid10.1145/3664647.3680560} and Diff-Unmix~\cite{diffunmix10656884}, further exploit diffusion priors and iterative denoising to improve perceptual quality. 
SFNet~\cite{SFNet}, FSNet~\cite{FSNet}, StarIR~\cite{starir11429607}, and MHNet~\cite{gao2025mixed}, explore frequency-domain modeling, linear attention, and hierarchical feature representations to enhance cross-task restoration. State space models have also been introduced into restoration. MambaIRV2~\cite{guo2025mambairv2}, ACL~\cite{aclgu2025acl}, for instance, improve spatial representation through multidirectional scanning and non-causal state modeling, while related studies investigate alternative scanning or hybrid designs to better capture spatial dependencies. 
PromptIR~\cite{potlapalli2023promptir} and CAPTNet~\cite{CAPTNet10526271} introduce degradation-aware prompts to guide restoration, while NDR~\cite{NDR10680296} learns neural degradation representations to capture shared degradation characteristics. BaryIR~\cite{baryir11417902} improves generalization by learning degradation-invariant representations through distribution alignment. More recently, large-model-based methods have incorporated vision-language and diffusion priors. VLU-Net~\cite{VLUNetZeng_2025_CVPR} uses vision-language representations to extract degradation-aware cues, AutoDIR~\cite{autodir10.1007/978-3-031-73661-2_19} combines vision-language guidance with latent diffusion priors, and Defusion~\cite{DefusionLuo_2025_CVPR} introduces degradation instruction diffusion for generalized restoration. Other methods, such as AdaIR~\cite{cui2025adair} and Perceive-IR~\cite{Perceive-IR10990319}, further improve robustness by disentangling degradation and content information or identifying fine-grained degradation severity.

These methods progressively extend image restoration from task-specific models to task-aligned and all-in-one frameworks. However, their restoration knowledge remains primarily encoded in fixed model parameters, and the inference process is generally determined by the representations learned during training. As a result, they cannot explicitly reuse newly observed action outcomes or adapt restoration strategies through accumulated execution experience. In this work, existing restoration networks are treated as callable tools, while the main focus is placed on how restoration experience can be structured, retrieved, and reused to support adaptive tool coordination.

\subsection{Image Restoration Agents}

Recent advances in large language models~\cite{LLMbrown2020language,LLMroziere2023code,LLMtouvron2023llama} and multimodal large language models~\cite{MLLMli2023blip,MLLMliu2023visual,MLLMsingh2025openai} have inspired agentic image restoration. Instead of treating restoration as a direct image-to-image mapping, restoration agents formulate it as a multi-step decision-making process, where the agent perceives degradations, plans operations, invokes tools, evaluates intermediate results, and revises the plan according to feedback.

Several representative systems have explored this direction. RestoreAgent~\cite{RestoreAgent} uses a vision-language model to generate restoration plans and call corresponding tools. AgenticIR~\cite{agenticir} abstracts human image processing into perception, scheduling, execution, reflection, and rescheduling, and introduces self-exploration to summarize restoration experience. MAIR~\cite{MAIRjiang2026multi} further improves efficiency by introducing a real-world degradation prior and a two-level multi-agent system, where a scheduler agent generates the global plan and expert agents handle specific degradations. HybridAgent~\cite{hybridagentli2026hybrid} proposes fast, slow, and feedback agents to adapt to different user prompts and highlights the error propagation problem caused by step-by-step single-degradation restoration. Restore-R1~\cite{lu2026restorer1} formulates tool selection as a reinforcement learning problem and trains a lightweight policy using multimodal perceptual feedback, reducing the need for iterative reflection and rollback.

Existing restoration agents improve flexibility by combining degradation perception, tool selection, execution, and feedback-based replanning. Nevertheless, most existing systems focus on selecting suitable tools or correcting an unsuccessful restoration trajectory. Their prior experience, when retained, is usually represented by static priors, prompt templates, or textual summaries, without explicitly modeling context-dependent action effects, evidence confidence, or memory consolidation. Causal-AgentIR addresses this limitation through a structured restoration memory graph and a memory curator that supports retrieval, reasoning, revision, reinforcement, consolidation, and selective forgetting.

\begin{figure*}[t]
\centering
\includegraphics[width=\linewidth]{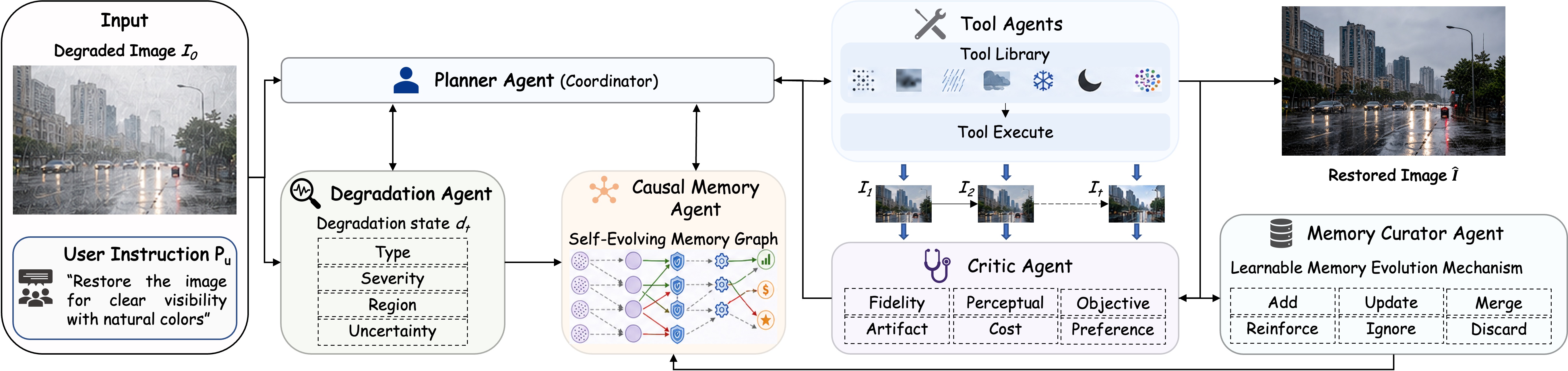}
\caption{Overview of the proposed Causal-AgentIR framework.}
\label{fig:framework}
\end{figure*}

\section{Method}
\label{sec:method}

\subsection{Overview}
\label{subsec:overview}

As shown in Figure~\ref{fig:framework}, given a degraded image $I_0$ and a user instruction $P_u$, Causal-AgentIR aims to generate a restored image $\hat{I}$  through hierarchical multi-agent collaboration. The overall framework consists of three coupled components. First, a hierarchical multi-agent system decomposes complex restoration into degradation analysis, restoration planning, tool expertise, causal memory reasoning, outcome critique, and memory curation. Second, a self-evolving causal memory graph stores structured restoration experience, including degradation patterns, image regions, restoration tools, actions, quality changes, computational costs, and user preferences. Third, a learnable memory evolution mechanism determines whether newly observed restoration experience should be added, updated, merged, reinforced, ignored, or discarded according to quality changes and feedback.

\subsection{Problem Formulation}
\label{subsec:problem_formulation}

Let $I_t$ denote the image state after the $t$-th restoration step. At each step, the system selects a restoration action
\begin{equation}
a_t=(T_i,\theta_t,m_t),
\end{equation}
where $T_i\in\mathcal{T}$ is the selected restoration tool, $\theta_t$ denotes its parameter configuration, and $m_t\in[0,1]^{H_I\times W_I}$ denotes the target region mask. Here, $H_I$ and $W_I$ denote the image height and width, respectively. For global restoration, $m_t=\mathbf{1}$. The restored image is obtained by
\begin{equation}
I_{t+1}
=
m_t\odot T_i(I_t;\theta_t)
+
(1-m_t)\odot I_t,
\end{equation}
where $\odot$ denotes element-wise multiplication. A restoration trajectory is denoted as
\begin{equation}
\tau=\{I_0,a_0,I_1,a_1,\ldots,a_{L-1},I_L\},
\end{equation}
where $L$ is the trajectory length. The objective is to maximize the cumulative restoration utility:
\begin{equation}
\max_{\pi}
\mathbb{E}_{\tau\sim\pi}
\left[
\sum_{t=0}^{L-1}\mathcal{R}_t
\right],
\end{equation}
where $\mathcal{R}_t$ denotes the restoration reward at step $t$, which is defined in Eq.~\eqref{eq:rewardr}.

In addition to restoration actions, Causal-AgentIR explicitly models memory evolution operations. At each step, the memory curator selects an operation
\begin{equation}
o_t\in
\{
\mathrm{add},
\mathrm{update},
\mathrm{merge},
\mathrm{reinforce},
\mathrm{ignore},
\mathrm{discard}
\}.
\end{equation}
The joint decision policy is formulated as
\begin{equation}
(a_t,o_t)\sim \pi(a,o|s_t,\mathcal{G}),
\end{equation}
where $s_t$ denotes the current restoration state and $\mathcal{G}$ is the causal memory graph.

To avoid ambiguity between the current observation and the retrieved memory, we first define the pre-retrieval state as
\begin{equation}
x_t=[\phi(I_t),d_t,q_t,h_t],
\end{equation}
where $\phi(I_t)$ is the visual representation of the current image, $d_t$ denotes the degradation state, $q_t$ denotes the quality state, and $h_t$ denotes the execution history. The causal memory agent retrieves a subgraph $g_t$ according to $x_t$ and $P_u$, and the final decision state is defined as
\begin{equation}
s_t=[x_t,g_t].
\end{equation}

\subsection{Hierarchical Multi-Agent System}
\label{subsec:hmas}

Causal-AgentIR decomposes complex restoration into a set of specialized agent roles, including a planner agent, a degradation agent, tool agents, a causal memory agent, a critic agent, and a memory curator agent. These agents collaborate in a closed loop of degradation analysis, causal memory retrieval, restoration planning, tool execution, outcome evaluation, and memory updating.

\textbf{Planner Agent.}
The planner agent serves as the global coordinator. It receives the user instruction, degradation analysis results, retrieved causal memory, and critique feedback, and then decomposes the restoration objective into executable subtasks. Specifically, it determines whether restoration should be performed globally or regionally, whether a single tool or a tool sequence should be used, and whether an all-in-one or mixed-degradation tool is preferable to step-by-step restoration.

\textbf{Degradation Agent.}
The degradation agent identifies global and local degradations from the input image. Its output includes degradation category, severity, spatial region, and uncertainty. 

\textbf{Causal Memory Agent.}
The causal memory agent retrieves relevant subgraphs from the causal memory graph and performs multi-hop causal reasoning. It estimates how a restoration tool or tool sequence may affect restoration quality under the current degradation condition. The retrieved causal evidence is provided to the planner, helping it avoid harmful operations and select more reliable restoration orders.

\textbf{Tool Agents.}
Tool agents are responsible for executing restoration operations in specific domains, including denoising, deblurring, deraining, dehazing, and desnowing. They may also include all-in-one or mixed-degradation restoration tools. After the planner generates the restoration plan, the corresponding tool agent executes the selected action.

\textbf{Critic Agent.}
The critic agent evaluates intermediate and final restoration results after each executed action. Its evaluation integrates fidelity-oriented metrics when paired references are available, no-reference perceptual metrics in real-world scenarios, artifact indicators, computational cost, and optional user or multimodal feedback. 

\textbf{Memory Curator Agent.}
The memory curator updates the causal memory according to observed restoration outcomes. It determines whether newly generated experience should be added as new knowledge, used to update an existing causal relation, merged with similar memories, reinforced as reliable knowledge, ignored as uninformative evidence, or discarded as unstable or harmful experience.

\subsection{Self-Evolving Causal Memory Graph}
\label{subsec:causal_memory}

The core of Causal-AgentIR is a self-evolving causal memory graph that stores restoration experience in a structured form. The graph is defined as
\begin{equation}
\mathcal{G}=(\mathcal{V},\mathcal{E}),
\end{equation}
where $\mathcal{V}$ and $\mathcal{E}$ denote the node set and edge set, respectively. The node set is defined as
\begin{equation}
\mathcal{V}
=
\{
V_d,V_r,V_T,V_a,V_q,V_c,V_u
\},
\end{equation}
where $V_d$, $V_r$, $V_T$, $V_a$, $V_q$, $V_c$, and $V_u$ denote degradation nodes, region nodes, tool nodes, action nodes, quality-change nodes, cost nodes, and user-preference nodes, respectively.

Each edge describes an empirical causal or conditional dependency between two nodes:
\begin{equation}
e_{ij}=(v_i,v_j,z_{ij},w_{ij},\gamma_{ij}),
\end{equation}
where $z_{ij}$ denotes the contextual condition, $w_{ij}$ denotes the estimated effect on restoration quality, and $\gamma_{ij}$ denotes the confidence score. In this paper, ``causal'' refers to an empirical interventional relation induced by actual restoration operations rather than a fully specified structural causal model. Specifically, under a given context, applying an action and observing the resulting quality change provides evidence for updating the corresponding causal relation. For example, an edge may encode that denoising before dehazing improves perceptual quality under strong noise and haze, or that deblurring before deraining may amplify rain streaks and reduce deraining effectiveness.

Given the pre-retrieval state $x_t$ and user instruction $P_u$, the causal memory agent constructs a query representation:
\begin{equation}
\psi_t=\Psi(x_t,P_u),
\end{equation}
where $\psi_t$ encodes degradation categories, severity levels, affected regions, user requirements, and execution history. A relevant subgraph is retrieved by
\begin{equation}
g_t=\mathrm{Retrieve}(\mathcal{G},\psi_t).
\end{equation}
For each edge $e$ in the graph, its retrieval relevance is computed as
\begin{equation}
\mathrm{Rel}(e|\psi_t)
=
\lambda_s\mathrm{Sim}(\psi_t,z_e)
+
\lambda_{\gamma}\gamma_e
+
\lambda_w |w_e|,
\end{equation}
where $\mathrm{Sim}(\cdot)$ measures contextual similarity, $\gamma_e$ is the edge confidence, and $w_e$ is the estimated causal effect. The term $|w_e|$ allows the agent to retrieve both beneficial and harmful high-impact relations, since negative evidence is also useful for avoiding unreliable restoration plans.

For a candidate restoration plan
\begin{equation}
\Pi_t=\{a_t,a_{t+1},\ldots,a_{t+H-1}\},
\end{equation}
with planning horizon $H$, the expected utility is estimated by aggregating related causal paths in the retrieved subgraph:
\begin{equation}
U(\Pi_t|s_t)
=
\sum_{\mathcal{P}\in\Omega(\Pi_t,g_t)}
\Gamma(\mathcal{P})
-
\lambda_C C(\Pi_t),
\end{equation}
where $\Omega(\Pi_t,g_t)$ denotes the set of causal paths related to $\Pi_t$, $C(\Pi_t)$ denotes computational cost, and $\Gamma(\mathcal{P})$ is the path-level causal score. For a path $\mathcal{P}=\{e_1,e_2,\ldots,e_J\}$, the score is defined as
\begin{equation}
\Gamma(\mathcal{P})
=
\left(\prod_{j=1}^{J}\gamma_{e_j}\right)
\left(\sum_{j=1}^{J}w_{e_j}\right).
\end{equation}

This formulation favors plans supported by reliable and positive causal evidence. Compared with unstructured textual summaries, the causal memory graph allows the agent to reason not only about which tool should be used, but also about why a specific operation or execution order is likely to be effective under the current degradation condition.

\subsection{Learnable Memory Evolution Mechanism}
\label{subsec:memory_evolution}

After executing action $a_t$, the critic and memory curator agents evaluate the transition from $I_t$ to $I_{t+1}$ and update the causal memory accordingly. The quality change is defined as
\begin{equation}
\Delta Q_t
=
\Phi(Q(I_{t+1}),Q(I_t),P_u),
\end{equation}
where $Q(\cdot)$ denotes the quality measurements and $\Phi(\cdot)$ aggregates them according to the user instruction. In practice, the quality change is computed as
\begin{equation}
\Delta Q_t
=
\Delta Q_t^{\mathrm{fid}}
+
\beta_1\Delta Q_t^{\mathrm{per}}
+
\beta_2\Delta Q_t^{\mathrm{obj}}
-
\beta_3\Delta Q_t^{\mathrm{art}}
-
\beta_4 C_t
+
\beta_5 R_t^{\mathrm{pref}},
\end{equation}
where $\Delta Q_t^{\mathrm{fid}}$ denotes fidelity improvement, $\Delta Q_t^{\mathrm{per}}$ denotes perceptual improvement, $\Delta Q_t^{\mathrm{obj}}$ denotes objective quality improvement, $\Delta Q_t^{\mathrm{art}}$ denotes artifact increase, $C_t$ denotes computational cost, and $R_t^{\mathrm{pref}}$ denotes preference alignment. In real-world scenarios without reference images, the fidelity term is omitted.

Each restoration transition is converted into an experience tuple:
\begin{equation}
\xi_t=(s_t,a_t,I_{t+1},\Delta Q_t,P_u,F_t),
\end{equation}
where $F_t$ denotes textual, multimodal, or user feedback. Given $\xi_t$ and the retrieved subgraph $g_t$, the curator selects a memory operation:
\begin{equation}
o_t=\pi_m(\xi_t,g_t),
\end{equation}
where
\begin{equation}
o_t\in
\{
\mathrm{add},
\mathrm{update},
\mathrm{merge},
\mathrm{reinforce},
\mathrm{ignore},
\mathrm{discard}
\}.
\end{equation}

\textbf{Add.}
If the observed experience is novel and produces a significant quality change, a new causal edge is added:
\begin{equation}
e_{\mathrm{new}}
=
(v_i,v_j,z_{\mathrm{new}},\Delta Q_t,\gamma_0),
\end{equation}
where $\gamma_0$ is the initial confidence score.

\textbf{Update.}
If the experience matches an existing edge $e_{ij}$, its causal effect is updated by
\begin{equation}
w_{ij}^{t+1}
=
\alpha w_{ij}^{t}
+
(1-\alpha)\Delta Q_t,
\end{equation}
where $\alpha$ is a momentum coefficient.

\textbf{Reinforce.}
If the new observation is consistent with the existing causal relation, the confidence score is increased:
\begin{equation}
\gamma_{ij}^{t+1}
=
\gamma_{ij}^{t}
+
\rho(1-\gamma_{ij}^{t}),
\end{equation}
where $\rho$ controls the reinforcement rate.

\textbf{Merge.}
If two memory entries describe similar causal relations under similar contexts, they are merged:
\begin{equation}
w_{\mathrm{merge}}
=
\frac{\gamma_1w_1+\gamma_2w_2}{\gamma_1+\gamma_2},
\end{equation}
\begin{equation}
\gamma_{\mathrm{merge}}
=
\max(\gamma_1,\gamma_2).
\end{equation}

\textbf{Ignore.}
If the observed experience has negligible quality change or low confidence, it is not used to update the graph:
\begin{equation}
\mathcal{G}^{t+1}=\mathcal{G}^{t},
\quad
\mathrm{if}
\quad
|\Delta Q_t|<\tau_Q
\ \mathrm{or}\
\delta_t>\tau_{\delta},
\end{equation}
where $\tau_Q$ and $\tau_{\delta}$ are thresholds for quality variation and uncertainty, respectively.

\textbf{Discard.}
If a memory entry repeatedly leads to negative outcomes or conflicts with new observations, it is weakened or removed:
\begin{equation}
e_{ij}\leftarrow \varnothing,
\quad
\mathrm{if}
\quad
\gamma_{ij}<\tau_{\gamma}
\ \mathrm{and}\
w_{ij}<-\tau_w.
\end{equation}

The memory-operation policy is trained to improve long-term restoration utility. Its reward is defined as
\begin{equation}
\label{eq:rewardr}
\mathcal{R}_t
=
\Delta Q_t
+
\eta_1 R_t^{\mathrm{mem}},
\end{equation}
where $R_t^{\mathrm{mem}}$ measures whether the updated memory improves future planning. It is estimated by
\begin{equation}
R_t^{\mathrm{mem}}
=
U(\Pi_{t+1}|\mathcal{G}^{t+1})
-
U(\Pi_{t+1}|\mathcal{G}^{t}),
\end{equation}
where $\mathcal{G}^{t}$ and $\mathcal{G}^{t+1}$ denote the memory graph before and after the update, respectively. Here, $\Pi_{t+1}$ denotes the candidate restoration plan generated from the next image state. 
Through this mechanism, Causal-AgentIR does not simply store all historical records. Instead, it selectively retains reliable causal knowledge, revises unstable relations, merges redundant experience, ignores uninformative transitions, and forgets misleading memory. This allows the agent to maintain compact, transferable, and continuously evolving restoration knowledge.

\section{Experiments}
\label{sec:exp}
In this section, we first present the experimental setup, followed by qualitative and quantitative comparison results. Finally, ablation studies are conducted to verify the effectiveness of the proposed method.

\subsection{Experimental Setup}
We evaluate our method under both all-in-one and task-specific settings.

\subsubsection{Datasets}
\textbf{Task-specific setting.} 
We evaluate the framework separately on image denoising, deblurring, deraining, dehazing, and desnowing. Unlike conventional task-specific methods that are explicitly provided with the degradation category, Causal-AgentIR receives only the degraded image and a generic restoration instruction unless otherwise specified. The degradation agent first estimates the degradation condition, after which the planner selects an appropriate restoration tool or tool sequence from the shared library.
\textbf{i)} Image deraining: we evaluate on several benchmark datasets, including Rain100H~\cite{Rain100}, Rain100L~\cite{Rain100}, Test100~\cite{Test100}, and Test1200~\cite{DIDMDN}.
\textbf{ii)} Image desnowing: we utilize the Snow100K~\cite{desnownet}, SRRS~\cite{JSTASRchen2020jstasr}, and CSD~\cite{HDCW-Netchen2021all} datasets. Following the protocol in~\cite{FSNet}, we randomly sample 2,000 images for evaluation.
\textbf{iii)} Image dehazing: we evaluate Causal-AgentIR on the daytime synthetic subsets of RESIDE~\cite{RESIDEli2018benchmarking}. 
\textbf{iv)} Image deblurring: we evaluate Causal-AgentIR on the GoPro dataset~\cite{Gopro} and HIDE dataset~\cite{HIDE}.
\textbf{v)} Image denoising: Evaluation is performed on CBSD68~\cite{BSD68}, Urban100~\cite{urban100}, and Kodak24~\cite{kodak}. Additive white Gaussian noise is generated with $\sigma \in {15,25,50}$.

\textbf{All-in-one setting.}
We evaluate on Test100~\cite{Test100} for deraining, Snow100K~\cite{desnownet} for desnowing, SOTS-Indoor~\cite{RESIDEli2018benchmarking} for dehazing, GoPro~\cite{Gopro} for deblurring, and CBSD68~\cite{BSD68} for denoising.

\subsubsection{Evaluation Metrics}

For paired datasets, we report peak signal-to-noise ratio (PSNR), structural similarity index measure (SSIM), and learned perceptual image patch similarity (LPIPS)~\cite{54zhang2018unreasonable}. 
For real-world images without reference images, we report Blind/Referenceless Image Spatial Quality Evaluator (BRISQUE)~\cite{59mittal2011blind},  Fog Aware Density Evaluator (FADE)~\cite{58choi2015referenceless}, and Neural Image Assessment (NIMA)~\cite{60talebi2018nima}. For PSNR, SSIM, and NIMA, higher values indicate better performance, whereas lower values are preferred for LPIPS, FADE, and BRISQUE. In all tables, the best and second-best results are highlighted in bold and underlined, respectively.

\begin{table}
\caption{Restoration tools used in the shared tool library.}
    \label{tab:restools}
    \centering
    \begin{tabular}{c|c}
    \hline
         Task& Tools
         \\
         \hline
         \hline
        Derain&  PPTformer~\cite{pptformerwang2025intra}, EfDeRain+~\cite{efderainguo2025efficientderain+}, MHNet~\cite{gao2025mixed}
        \\
        Desnow & MSP-Former~\cite{mspformer10095605}, PW-FNet~\cite{pwfnet11433521},PEUNet~\cite{PEUNet10830558}
        \\
        Dehaze & PGH$^2$Net~\cite{PGH2Netisu2025prior},  DEA-Net-CR~\cite{deanetchen2024dea}, ECFNet~\cite{gao2026emphasizing}
        \\
        Deblur &  Omni-Deblurring~\cite{10919160},ALGNet~\cite{ALGgao2024learning}, MDT~\cite{MDTChen_2025_CVPR} 
        \\
        Denoise & FSNet~\cite{FSNet}, StarIR~\cite{starir11429607},  Restormer~\cite{Zamir2021Restormer}
        \\
        All-in-one & Perceive-IR~\cite{Perceive-IR10990319}, BaryIR~\cite{baryir11417902}, Defusion~\cite{DefusionLuo_2025_CVPR}
         \\
         \hline
    \end{tabular}
\end{table}

\begin{table*}
\centering
\caption{Quantitative results in the all-in-one setting with restoration models and restoration agents. Denoising results are reported for the noise level 15. }
\label{tb:allrshbn}
    \resizebox{\linewidth}{!}{
\begin{tabular}{c|c|cccccccccc|cc}
    \hline
   \multirow{2}{*}{Type} & \multirow{2}{*}{Methods} & \multicolumn{2}{c}{Deraining} & \multicolumn{2}{c}{Desnowing} & \multicolumn{2}{c}{Dehazing}& \multicolumn{2}{c}{Deblurring}& \multicolumn{2}{c|}{Denoising}  & \multicolumn{2}{c}{Average} 
    \\
   & &PSNR $\uparrow$ &  SSIM $\uparrow$  &PSNR $\uparrow$ &SSIM $\uparrow$ & PSNR $\uparrow$&SSIM $\uparrow$ &PSNR $\uparrow$ & SSIM $\uparrow$&PSNR $\uparrow$ & SSIM $\uparrow$&PSNR $\uparrow$ & SSIM $\uparrow$
    \\
    \hline\hline
      \multirow{6}{*}{\rotatebox{90}{Models}}
      &Perceive-IR~\cite{Perceive-IR10990319}& 30.93 & 0.902 & 32.99 & 0.948 & 40.05 & \underline{0.996} & 29.79 & 0.889 & 33.81 & 0.962 & 33.51 & 0.940
        \\
        &Defusion~\cite{DefusionLuo_2025_CVPR}& 30.66 & 0.905 & 33.33 & 0.952 & 40.21 & 0.995& 29.98 & 0.889 & 33.73& 0.962 & 33.58 & 0.940
        \\
        &AllRestorer~\cite{Allrestor11367271} &30.71 &0.903 & 33.17& 0.951 & 40.11 &0.991 & 29.63 &0.882& 33.62 & 0.959 & 33.45 & 0.937
        \\
        &BaryIR~\cite{baryir11417902} & 30.82 &0.907 & 33.05& 0.948& 40.20& 0.995 & 29.52& 0.881& 33.55 & 0.963 & 33.43 & 0.939
    \\
    \hline
 \multirow{6}{*}{\rotatebox{90}{Agents}}
&MAIR~\cite{MAIRjiang2026multi}
& 31.23 & 0.921 & 33.97 & 0.957 & 41.48 & \underline{0.996} & 34.08 & 0.965 & 34.02 & \textbf{0.971 }&34.96 &0.962
\\
&AgenticIR~\cite{agenticir}
& 31.10 & 0.920 & 34.17 & 0.961 & 41.25 & \underline{0.996} & 34.18 & 0.963 & 34.25 & \underline{0.969} &34.99  &0.962 
\\
&HybridAgent~\cite{hybridagentli2026hybrid}
& \underline{31.56} & \textbf{0.923} & 34.03 & 0.959 & \underline{41.52} & \underline{0.996} & 34.29 & 0.968 & 34.28 & 0.968 & 35.17 &\underline{0.963}
\\
&IAMAgent~\cite{wei2026iamagent}
& 31.55 & \underline{0.922} & \underline{34.50} & \underline{0.962} & 41.44 & 0.996 & \underline{34.31} &\textbf{0.974} & \textbf{34.40} & 0.965 &\underline{35.24}  &\textbf{0.964}
\\
&Restore-R1~\cite{lu2026restorer1}
& 31.48 & 0.920 & 34.46 & \underline{0.962} & 41.39 & \textbf{0.997} & 34.15 & 0.968 & 34.17 & 0.968 &35.13  &\underline{0.963}
\\
&\cellcolor{gray!20}\textbf{Causal-AgentIR(Ours)}
&\cellcolor{gray!20}\textbf{31.89}
&\cellcolor{gray!20}\textbf{0.923}
&\cellcolor{gray!20}\textbf{34.85}
&\cellcolor{gray!20}\textbf{0.965}
&\cellcolor{gray!20}\textbf{41.88}
&\cellcolor{gray!20}\textbf{0.997}
&\cellcolor{gray!20}\textbf{34.73}
&\cellcolor{gray!20}\underline{0.970}
&\cellcolor{gray!20}\underline{34.37}
&\cellcolor{gray!20}0.965
&\cellcolor{gray!20}\textbf{35.55}
&\cellcolor{gray!20}\textbf{0.964}
\\
    \hline
\end{tabular}}
\end{table*}

\subsubsection{Training Details}

We follow the same implementation setting as MAIR~\cite{MAIRjiang2026multi}. Specifically, the DepictQA model adopted in AgenticIR~\cite{agenticir} and GPT-4o are used for planning, degradation identification, and critique. The restoration tool library is reported in Table~\ref{tab:restools}.
All tools use publicly released checkpoints whenever available. Otherwise, they are retrained following the settings reported in the corresponding papers. The parameters of all restoration tools are frozen during agent training. Consequently, the improvements achieved by Causal-AgentIR mainly arise from degradation analysis, tool selection, execution ordering, regional application, and memory-guided planning rather than from modifying the restoration networks themselves. To ensure a fair comparison among agentic methods, all agents are given access to the same restoration tool library.

\subsection{All-in-one Setting Results}

Table~\ref{tb:allrshbn} reports the quantitative comparison between restoration models and restoration agents under the all-in-one R+S+H+B+N setting. Overall, restoration agents consistently outperform model-based all-in-one methods across most degradation types. The best restoration method, Defusion~\cite{DefusionLuo_2025_CVPR}, achieves an average PSNR of 33.58 dB. In contrast, even the weakest restoration agent in terms of average PSNR, MAIR~\cite{MAIRjiang2026multi}, reaches 34.96 dB, surpassing the best restoration model by 1.38 dB in PSNR. This clear performance gap indicates that agentic restoration is more effective than directly applying a single unified restoration model, especially when diverse degradation types need to be handled within the same inference framework.

\textbf{Comparison with restoration models.}
Recent all-in-one restoration models achieve competitive results by learning unified representations for multiple degradation types. Among them, Defusion~\cite{DefusionLuo_2025_CVPR} obtains the strongest average PSNR of 33.58 dB. In comparison, Causal-AgentIR improves the average PSNR from 33.58 dB to 35.55 dB. This improvement indicates that, beyond learning a single unified mapping, coordinating multiple restoration tools with degradation-aware planning and memory-guided decision-making can better adapt to heterogeneous restoration requirements.

\textbf{Comparison with existing restoration agents.}
Compared with existing restoration agents, Causal-AgentIR also achieves the best overall results. The strongest competing agent is IAMAgent~\cite{wei2026iamagent}, which obtains an average PSNR of 35.24 dB. Causal-AgentIR further improves the average PSNR to 35.55 dB, corresponding to gains of 0.31 dB. These results suggest that the proposed self-evolving causal memory provides more effective guidance for tool selection and execution ordering than conventional agentic restoration pipelines that mainly rely on tool search and trajectory-level reflection.

\textbf{Performance across individual degradation types.}
Causal-AgentIR achieves the best PSNR on four out of five tasks, including deraining, desnowing, dehazing, and deblurring. For denoising, Causal-AgentIR obtains the second-best PSNR of 34.37 dB, slightly lower than IAMAgent but still higher than the other agentic and model-based methods.

\subsection{Task-specific Setting Results}
To demonstrate that the proposed Causal-AgentIR is effective not only for all-in-one restoration but also for task-specific scenarios, we conduct experiments on five representative image restoration tasks: image deraining, image desnowing, image dehazing, image deblurring and image denoising.

\begin{table*}
\centering
\caption{Image deraining results in the task-specific setting.}
\label{tb:derain}
    \resizebox{\linewidth}{!}{
\begin{tabular}{c|cccccccc|cc}
    \hline
    \multicolumn{1}{c|}{} & \multicolumn{2}{c}{Test100}  & \multicolumn{2}{c}{Test1200} & \multicolumn{2}{c}{Rain100H} & \multicolumn{2}{c|}{Rain100L} & \multicolumn{2}{c}{Average} 
    \\
   Methods &PSNR $\uparrow$ &  SSIM $\uparrow$  & PSNR $\uparrow$ & SSIM $\uparrow$ &PSNR $\uparrow$ &SSIM $\uparrow$ & PSNR $\uparrow$&SSIM $\uparrow$ &PSNR $\uparrow$ & SSIM $\uparrow$
    \\
    \hline\hline
     EfDeRain+~\cite{efderainguo2025efficientderain+} &31.10 &0.911 &33.12 & 0.925 & \textbf{34.57} &\textbf{0.957} & 39.03 & 0.972 &34.46 & 0.941
     \\
     MAIR~\cite{MAIRjiang2026multi}
& 31.23 & 0.921 & 33.28 & 0.927 & 34.40 & 0.955 & 39.99 & \underline{0.984} & 34.73 & 0.947
\\
AgenticIR~\cite{agenticir}
& 31.10 & 0.920 & 33.31 & \underline{0.928} & 34.52 & \underline{0.956} & 39.71 & 0.980 & 34.66 & 0.946
\\
HybridAgent~\cite{hybridagentli2026hybrid}
& \underline{31.56} & \textbf{0.923} &  33.39 &0.926 & 34.28 &\underline{0.956} & 39.72 & 0.981 & 34.74 & \underline{0.947}
\\
IAMAgent~\cite{wei2026iamagent}
& 31.55 & \underline{0.922} & \underline{33.43} & 0.925 & \underline{34.55} & 0.951 & 39.41 & 0.979 & 34.74 & 0.944
\\
Restore-R1~\cite{lu2026restorer1}
& 31.48 & 0.920 & 33.25 & 0.926 & 34.27 & 0.955 & \underline{40.00} & 0.983 & \underline{34.75} &0.946
\\
      \hline
    \rowcolor{gray!20}  \textbf{Causal-AgentIR(Ours)}  & \textbf{31.89}	&\textbf{0.923}	&\textbf{33.52}	&\textbf{0.929}	&34.53	&0.954	&\textbf{40.08}	&\textbf{0.985}	&\textbf{35.01}	&\textbf{0.948}
    \\
    \hline
\end{tabular}}
\end{table*}

\subsubsection{Image Deraining}

Following the evaluation protocol in prior work~\cite{gao2025mixed}, we report PSNR and SSIM on the Y channel in the YCbCr color space for image deraining. Table~\ref{tb:derain} summarizes the quantitative results under the task-specific setting.
Overall, Causal-AgentIR achieves the best average performance among all compared methods, with 35.01 dB PSNR. Compared with the strongest existing agent Restore-R1~\cite{lu2026restorer1}, Causal-AgentIR brings a gain of 0.26 dB. 
Compared with restoration models, restoration agents generally achieve better results. For example, the strongest baseline, EfDeRain+~\cite{efderainguo2025efficientderain+}, obtains 34.46 dB PSNR and 0.941 SSIM on average, while existing agents improve the average PSNR to 34.66--34.75 dB. Causal-AgentIR further increases the average PSNR to 35.01 dB, demonstrating that agentic restoration can better exploit tool coordination and degradation-aware decision-making than directly applying a single restoration model.
On individual datasets, Causal-AgentIR achieves the best PSNR on Test100, Test1200, and Rain100L. On Rain100H, EfDeRain+ obtains the highest PSNR, while Causal-AgentIR remains highly competitive, achieving 34.53 dB.

\begin{table}
    \centering
        \caption{Image  desnowing results in the task-specific setting.}
    \label{tab:snow}
    \resizebox{\linewidth}{!}{
    \begin{tabular}{c|cccccc}
    \hline
    \multicolumn{1}{c|}{} & \multicolumn{2}{c}{CSD}  & \multicolumn{2}{c}{SRRS} & \multicolumn{2}{c}{Snow100K}
    \\
   Methods & PSNR $\uparrow$ & SSIM $\uparrow$  & PSNR $\uparrow$ & SSIM $\uparrow$ & PSNR $\uparrow$ & SSIM $\uparrow$
   \\
   \hline
   \hline
PEUNet~\cite{PEUNet10830558}&37.28 &0.97 &31.89 &\textbf{0.98} &34.11 &\underline{0.96}
\\
PW-FNet~\cite{pwfnet11433521}&-&-&32.74&\textbf{0.98}&\underline{34.50}&0.95
\\
 StarIR~\cite{starir11429607}  & \textbf{38.57} & \textbf{0.99} & \underline{32.75} & \textbf{0.98} & 34.46 & \underline{0.96}
\\
 MAIR~\cite{MAIRjiang2026multi}
&38.41  &\underline{0.98}  &32.70 &\underline{0.97}   & 33.97 &\underline{0.96}
\\
AgenticIR~\cite{agenticir}
&38.55  &\underline{0.98}  &32.73 &\underline{0.97} & 34.17 & \underline{0.96}
\\
HybridAgent~\cite{hybridagentli2026hybrid}
&38.29  &0.97  &32.68 &0.96 & 34.03 & \underline{0.96}
\\
IAMAgent~\cite{wei2026iamagent}
&38.38  &\underline{0.98}  &32.74 &\textbf{0.98} & \underline{34.50} & \underline{0.96}
\\
Restore-R1~\cite{lu2026restorer1}
&38.49  &0.97  &32.66 &\underline{0.97}  &34.46  &\underline{0.96}
\\
\hline
\rowcolor{gray!20} \textbf{Causal-AgentIR(Ours)} &\underline{38.56}	&\underline{0.98}  &\textbf{32.89}	&\textbf{0.98}		&\textbf{34.85}&\textbf{0.97}

         \\
         \hline
    \end{tabular}}
\end{table}

\subsubsection{Image Desnowing}

Table~\ref{tab:snow} reports quantitative comparisons for image desnowing on three benchmark datasets, including CSD, SRRS, and Snow100K. Overall, Causal-AgentIR achieves the best or second-best performance across all datasets, showing strong robustness under different snow degradation conditions.
On the CSD dataset~\cite{HDCW-Netchen2021all}, StarIR~\cite{starir11429607} obtains the highest PSNR and SSIM. Causal-AgentIR achieves a very close result, ranking second and nearly matching the best-performing baseline. 
On SRRS~\cite{JSTASRchen2020jstasr}, Causal-AgentIR achieves the best PSNR of 32.89 dB and the best SSIM of 0.98. Compared with the strongest existing result in PSNR, obtained by StarIR with 32.75 dB, Causal-AgentIR brings a gain of 0.14 dB. On Snow100K~\cite{desnownet}, Causal-AgentIR also achieves the best performance.
These results show that Causal-AgentIR provides consistent gains on SRRS and Snow100K while maintaining near-best performance on CSD.

\begin{table}
    \centering
        \caption{Image dehazing results in the task-specific setting.}
    \label{tab:sot}
    \resizebox{\linewidth}{!}{
    \begin{tabular}{c|cccc}
    \hline
    \multicolumn{1}{c|}{} & \multicolumn{2}{c}{SOTS-Indoor}  & \multicolumn{2}{c}{SOTS-Outdoor} 
    \\
   Methods & PSNR $\uparrow$ & SSIM $\uparrow$  & PSNR $\uparrow$ & SSIM $\uparrow$ 
   \\
   \hline
   \hline
PGH$^2$Net~\cite{PGH2Netisu2025prior}&\underline{41.70} &\underline{0.996}&37.52 &0.989
\\
 MAIR~\cite{MAIRjiang2026multi}
&41.48  &\underline{0.996}  &39.03 & 0.993
\\
AgenticIR~\cite{agenticir}
&41.25  &\underline{0.996}  &39.17 & 0.994
\\
HybridAgent~\cite{hybridagentli2026hybrid}
&41.52  &0.996  &39.09 &\textbf{0.996}
\\
IAMAgent~\cite{wei2026iamagent}
&41.44  &\underline{0.996}  &39.17 &\textbf{0.996} 
\\
Restore-R1~\cite{lu2026restorer1}
&41.39  &\textbf{0.997}  &39.11 &\underline{0.995} 
\\
\hline
\rowcolor{gray!20} \textbf{Causal-AgentIR(Ours)} &\textbf{41.88}	&\textbf{0.997}  &\textbf{39.22}	&\textbf{0.996}		
         \\
         \hline
    \end{tabular}}
\end{table}

\begin{table}
\centering
\caption{Image deblurring results in the task-specific setting. \label{tb:deblurgh}}

\begin{tabular}{ccccc}
    \hline
    \multicolumn{1}{c}{} & \multicolumn{2}{c}{GoPro}  & \multicolumn{2}{c}{HIDE} 
    \\
   Methods & PSNR $\uparrow$ & SSIM $\uparrow$ & PSNR $\uparrow$ & SSIM $\uparrow$   
    \\
    \hline\hline
    StarIR~\cite{starir11429607} & \underline{34.34}&\underline{0.970} &\underline{32.12}&\underline{0.951}
    \\
     MAIR~\cite{MAIRjiang2026multi}
&34.08  &0.965  &32.06 & 0.950
\\
AgenticIR~\cite{agenticir}
&34.18  &0.963  &\underline{32.12} & 0.949
\\
HybridAgent~\cite{hybridagentli2026hybrid}
&34.29  &0.968  &32.04 &0.950
\\
IAMAgent~\cite{wei2026iamagent}
&34.31  &\textbf{0.974}  &32.09 &\textbf{0.952} 
\\
Restore-R1~\cite{lu2026restorer1}
&34.15  &0.968  &32.11 &\underline{0.951} 
\\
\hline
\rowcolor{gray!20} \textbf{Causal-AgentIR(Ours)} &\textbf{34.73}	&\underline{0.970}  &\textbf{32.20}	&\textbf{0.952}		
         \\
    \hline
\end{tabular}
\end{table}

\subsubsection{Image Dehazing}
Table~\ref{tab:sot} reports quantitative comparisons on the SOTS~\cite{RESIDEli2018benchmarking} benchmark. Causal-AgentIR achieves the best performance on both SOTS-Indoor and SOTS-Outdoor. On SOTS-Indoor, it obtains 41.88 dB PSNR, outperforming the strongest  baseline PGH$^2$Net~\cite{PGH2Netisu2025prior} by 0.18 dB in PSNR. On SOTS-Outdoor, Causal-AgentIR reaches 39.22 dB PSNR, surpassing existing restoration agents such as AgenticIR~\cite{agenticir} and IAMAgent~\cite{wei2026iamagent}.

\begin{figure*} 
    \centerline{\includegraphics[width=1\linewidth]{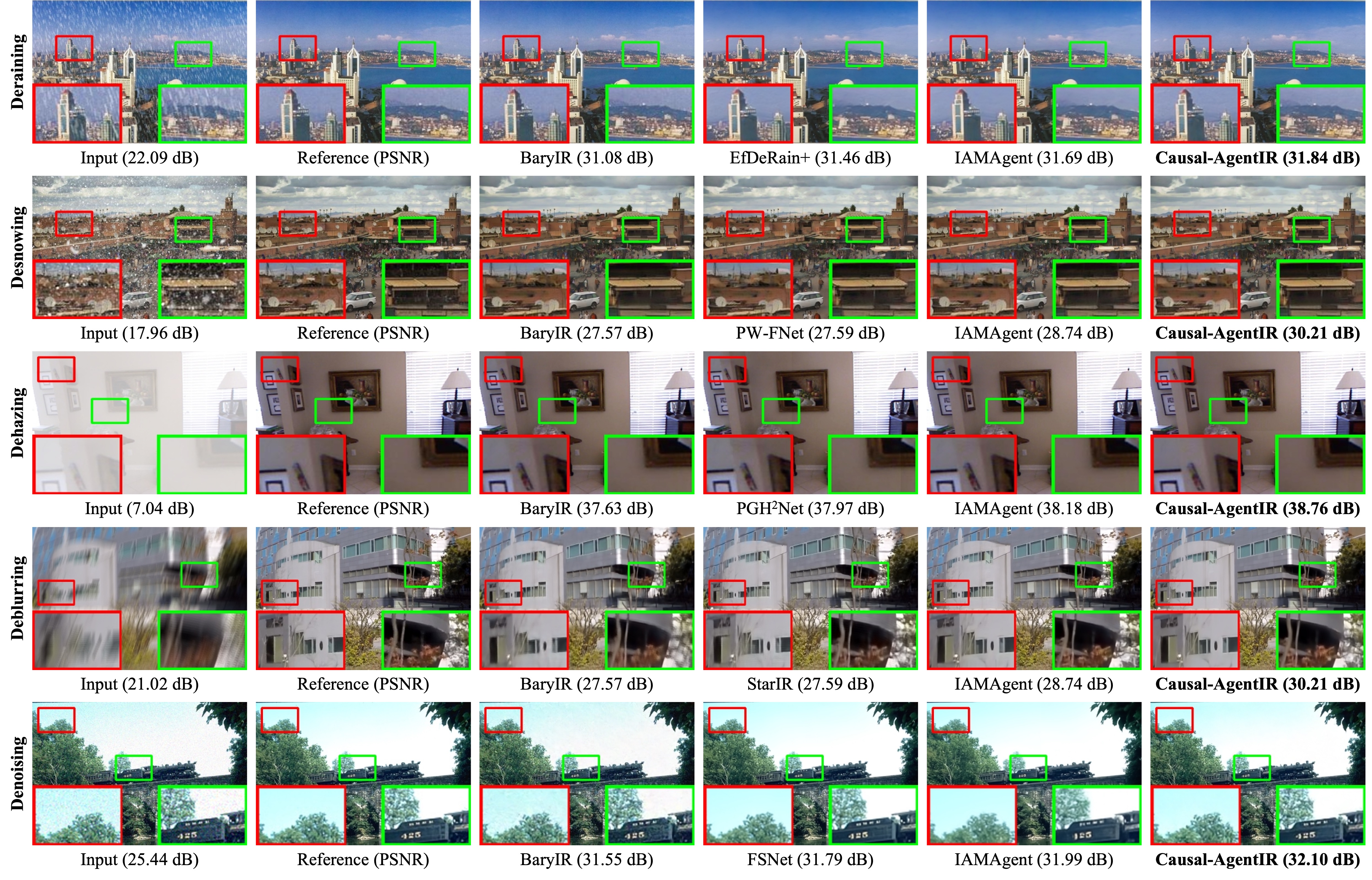}}
	\caption{Qualitative comparison of different restoration methods. The fourth column presents the results of the corresponding task-specific model, while the remaining columns show results obtained under the all-in-one setting.}
 \label{fig:all-in-one}
\end{figure*}

\subsubsection{Image Deblurring}

Table~\ref{tb:deblurgh} reports quantitative comparisons on the GoPro~\cite{Gopro} and HIDE~\cite{HIDE} datasets. Causal-AgentIR achieves the best PSNR on both benchmarks. On GoPro, it obtains 34.73 dB PSNR, outperforming the strongest competing method StarIR~\cite{starir11429607} by 0.39 dB. 
On HIDE, Causal-AgentIR achieves 32.20 dB PSNR, surpassing the second-best PSNR of 32.12 dB achieved by StarIR~\cite{starir11429607} and AgenticIR~\cite{agenticir}.

\begin{table}
    \centering
    \caption{Image denoising results in the task-specific setting.}
    \label{tab:denoisec}
    \resizebox{\linewidth}{!}{
    \begin{tabular}{c|ccccccccc}
    \hline
      & \multicolumn{3}{c}{CBSD68}  & \multicolumn{3}{c}{Kodak24} & \multicolumn{3}{c}{Urban100} 
    \\
   Methods  & 15  & 25 & 50 & 15  & 25 & 50 & 15  & 25 & 50
   \\
   \hline
   \hline
SFNet~\cite{SFNet} & 34.09 & 31.49 & 28.02 & 34.93 & 32.42 & 29.25&34.19 &32.01 & 29.03
\\
FSNet~\cite{FSNet} & 34.11 & 31.51 & 28.01 & \underline{34.95} & 32.42 & \underline{29.27}&34.15 & 32.04 & 29.15
\\
Perceive-IR~\cite{Perceive-IR10990319} &\underline{34.38} &\textbf{31.74} &28.53 &34.84 &\underline{32.50} &29.16 &34.86 &32.55 &29.42
\\
VLU-Net~\cite{VLUNetZeng_2025_CVPR}& 34.35 &\underline{31.72} &28.46 &- &- &-&\underline{34.92} &\textbf{32.71} &\textbf{29.61}
\\
     MAIR~\cite{MAIRjiang2026multi}
&34.02  &31.69 & 28.51 & 34.91 & 32.45 & 29.22 & 34.87 & \underline{32.70} & 29.49
\\
AgenticIR~\cite{agenticir}
&34.25  & 31.71 & 28.50 & 34.89 & 32.44 & 29.25 & 34.83 & 32.68 & 29.46 
\\
HybridAgent~\cite{hybridagentli2026hybrid}
&34.28  & 31.66 & 28.43 & 34.93 & \underline{32.50} & 29.23 & 34.91 & 32.68 & \underline{29.55}
\\
IAMAgent~\cite{wei2026iamagent}
&\textbf{34.40}  & \textbf{31.74} & \underline{28.65} & 34.94 & 32.41 & 29.26 & 34.90 & 32.66 & 29.49
\\
Restore-R1~\cite{lu2026restorer1}
&34.17  & 31.70 & 28.44 & 34.88 & 32.47 & 29.21 & 34.90 & 32.64 & 29.47
\\
\hline
\rowcolor{gray!20} \textbf{Causal-AgentIR(Ours)} &34.37	&\textbf{31.74} &\textbf{28.72} &\textbf{34.99}	&\textbf{32.52}	& \textbf{29.31} & \textbf{34.98} & 32.56 & \textbf{29.61}
         \\
    \hline
    \end{tabular}}
\end{table}

\begin{table}
\centering
\caption{Real-world results.}
\label{tb:allrealw}
\resizebox{\linewidth}{!}{
\begin{tabular}{c|ccc|ccc|cc}
    \hline
  \multirow{2}{*}{Methods} & \multicolumn{3}{c|}{RealRain-1k-L} & \multicolumn{3}{c|}{RTTS} & \multicolumn{2}{c}{SIDD}
    \\
    &PSNR $\uparrow$ &  SSIM $\uparrow$ &LPIPS $\downarrow$ &FADE $\downarrow$ &BRISQUE $\downarrow$ &NIMA $\uparrow$&PSNR $\uparrow$ & SSIM $\uparrow$
    \\
    \hline\hline
  ECFNet~\cite{gao2026emphasizing} & 23.69 & 0.756 & 0.401 & 1.394 & 29.105 &4.372 & 24.33 & 0.471 
    \\
      Perceive-IR~\cite{Perceive-IR10990319}& 27.31 & 0.901  & 0.372 & 1.264 & 23.694 &4.613 & 24.53 & 0.497
        \\
        MAIR~\cite{MAIRjiang2026multi}
&27.53 & 0.907 & \underline{0.335} & \underline{1.190} & \underline{23.231} & 4.874 & 24.68 & 0.500
\\
HybridAgent~\cite{hybridagentli2026hybrid}
&\underline{27.65} &0.909 &0.351 & 1.198 & 23.280 & 4.811 & \underline{24.70} & \underline{0.509}
\\
IAMAgent~\cite{wei2026iamagent}
&27.61 & \underline{0.912} & 0.344 & 1.203 & 23.284 & \underline{4.887} & 24.66 & 0.498
\\
\hline
      \cellcolor{gray!20}\textbf{Causal-AgentIR(Ours)} & \cellcolor{gray!20}\textbf{27.78} 
      &\cellcolor{gray!20}\textbf{0.913}
      & \cellcolor{gray!20}\textbf{0.332}
      & \cellcolor{gray!20}\textbf{1.145} 
      & \cellcolor{gray!20}\textbf{23.178}
      & \cellcolor{gray!20}\textbf{4.892} 
      & \cellcolor{gray!20}\textbf{24.71}
      &\cellcolor{gray!20}\textbf{0.513}
    \\
    \hline
\end{tabular}}
\end{table}

\subsubsection{Image Denoising}

Table~\ref{tab:denoisec} reports quantitative denoising results under Gaussian noise levels $\sigma=15,25,50$ on CBSD68, Kodak24, and Urban100. Overall, Causal-AgentIR achieves competitive or best performance across most settings, demonstrating stable denoising ability under different noise intensities and image contents.
On CBSD68, Causal-AgentIR obtains the best result at $\sigma=50$ with 28.72 dB and achieves the best tied performance at $\sigma=25$ with 31.74 dB. At $\sigma=15$, it reaches 34.37 dB, which is close to the best result of IAMAgent~\cite{wei2026iamagent}. On Kodak24, Causal-AgentIR achieves the best PSNR under all three noise levels, reaching 34.99 dB, 32.52 dB, and 29.31 dB, respectively. 
On Urban100, which contains rich repetitive structures and high-frequency details, Causal-AgentIR obtains the best PSNR at $\sigma=15$ and $\sigma=50$.

\subsection{Qualitative Comparison}
\label{subsec:qualitative}

Figure~\ref{fig:all-in-one} presents qualitative comparisons under different degradation scenarios. The fourth column shows the restoration results produced by the corresponding task-specific model, while the remaining compared methods are evaluated under the all-in-one setting. Compared with existing all-in-one restoration models and restoration agents, Causal-AgentIR produces visually cleaner results and recovers finer image details.
Specifically, existing all-in-one methods can remove dominant degradations to some extent, but they often suffer from residual artifacts, over-smoothed textures, or incomplete restoration in locally degraded regions. Although task-specific models achieve strong performance on their corresponding degradation types, their applicability is limited when the input contains other degradations. In contrast, Causal-AgentIR adaptively analyzes degradation conditions, retrieves relevant causal memory, and selects suitable restoration operations through multi-agent collaboration. As a result, it better preserves structural edges, local textures, and natural image appearance while suppressing degradation artifacts more effectively.
These qualitative results are consistent with the quantitative comparisons and further demonstrate that Causal-AgentIR improves not only numerical restoration accuracy but also perceptual visual quality in diverse restoration scenarios.

\subsection{Generalization}
\subsubsection{Zero-shot generalization in real-world scenes}
We further evaluate the zero-shot generalization capability of different methods under the unified R+S+H+B+N setting on real-world benchmarks, including RealRain-1k-L~\cite{realrain1li2022toward}, RTTS~\cite{Rttsli2018benchmarking}, and SIDD~\cite{ssidabdelhamed2018high}. Table~\ref{tb:allrealw} summarizes the quantitative results.
On RealRain-1k-L, Causal-AgentIR achieves the best performance across all three metrics. Compared with the strongest competing method HybridAgent~\cite{hybridagentli2026hybrid}, Causal-AgentIR improves PSNR by 0.13 dB. It also reduces LPIPS from 0.335, achieved by MAIR~\cite{MAIRjiang2026multi}, to 0.332, indicating better perceptual restoration quality under real-world rain degradation.
On RTTS, Causal-AgentIR also obtains the best results, achieving the lowest FADE score of 1.145, the lowest BRISQUE score of 23.178, and the highest NIMA score of 4.892. These results suggest that Causal-AgentIR improves haze removal while maintaining more natural image statistics and perceptual quality.
On SIDD, Causal-AgentIR also achieves the best performance. Overall, Causal-AgentIR achieves the best or highly competitive performance across real-world rain, haze, and noise benchmarks, demonstrating strong zero-shot generalization to unseen real-world degradations. This improvement can be attributed to degradation-aware tool coordination and causal memory-guided planning, which enable the agent to adapt restoration decisions to complex real-world degradation conditions.

\begin{table*}
\centering
\caption{Quantitative results of different models conducted on various combinations of degradation types.}
\label{tb:mix}
\resizebox{\linewidth}{!}{
\begin{tabular}{ccccccccccccccc}
    \hline
   Methods& H + R + N  & H + R &H + N& R + N & H &R&N
    \\
    \hline\hline
Perceive-IR~\cite{Perceive-IR10990319}&30.22/0.894	&31.55/0.945	&28.32/0.871	&30.52/0.849	&31.90/0.980	&38.95/0.983	&31.88/0.923	
\\
DSwinIR~\cite{DswinIR11304568} & 30.09/0.899 & 31.07/0.960 & 28.45/0.873 & 31.29/0.915 & 31.77/0.973 & 39.09/0.982 & 32.22/0.923 
\\
 MAIR~\cite{MAIRjiang2026multi}& 31.25/0.912 & 31.56/0.973 & 31.33/0.907 & 32.07/0.919 & \underline{39.43}/\textbf{0.993 }& \underline{40.17/0.992} &32.76/0.935
\\
AgenticIR~\cite{agenticir}& 31.31/0.912 & 31.62/0.971 & 31.44/0.910 & 32.15/\underline{0.921} & \underline{39.43/0.991} & 40.06/0.991 &\underline{32.83/0.937}
\\
HybridAgent~\cite{hybridagentli2026hybrid}& 31.22/0.912 & \underline{31.88}/\textbf{0.976}& \underline{31.52/0.911} & \underline{32.31}/\textbf{0.922} & 39.39/\textbf{0.993} & 39.99/0.990 &32.54/0.933
\\
IAMAgent~\cite{wei2026iamagent}& \underline{31.52/0.926} & 31.70/\underline{0.975} & 31.09/0.905 & 31.98/0.920 & 39.21/0.990 & 40.11/\underline{0.992} &32.58/0.933
\\
Restore-R1~\cite{lu2026restorer1}& 31.20/0.910 & 31.73/\textbf{0.976} & 31.27/0.910 & 32.22/0.918 & 39.08/\underline{0.991} & 39.96/0.989 &32.81/0.929
\\
\hline
\rowcolor{gray!20}\textbf{Causal-AgentIR(Ours)} &  \textbf{31.66/0.927} & \textbf{32.03/0.976} & \textbf{31.80/0.912} & \textbf{32.49}/0.920 & \textbf{39.55}/\underline{0.991} & \textbf{40.22/0.993} &\textbf{33.01/0.942}

    \\
    \hline
\end{tabular}}
\end{table*}

\begin{figure*} 
    \centerline{\includegraphics[width=1\linewidth]{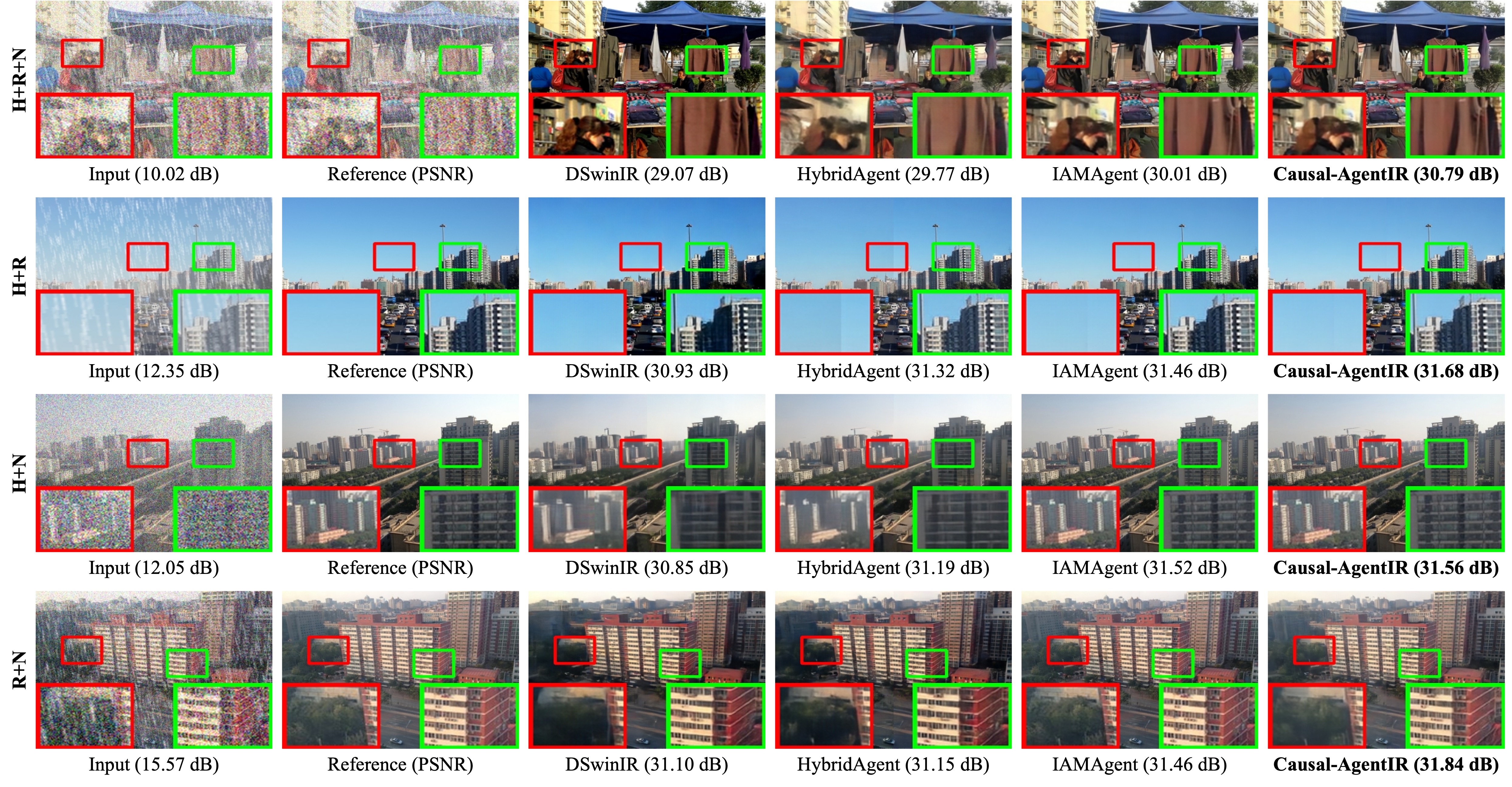}}
	\caption{Multi-degradation image restoration results across various degradation combinations.}
 \label{fig:mixm}
\end{figure*}

\subsubsection{Generalization to Mixed Degradations}
To further evaluate the robustness of different methods under complex real-world conditions, we conduct experiments on mixed degradation scenarios, where haze, rain, and noise appear either individually or in combination. The quantitative results are reported in Table~\ref{tb:mix}, and qualitative comparisons are shown in Figure~\ref{fig:mixm}.
Overall, restoration agents clearly outperform restoration models under most degradation settings. Model-based methods, such as FDTANet~\cite{FDTANetgao2025frequency}, Ref-IRT~\cite{REF-IRT}, Perceive-IR~\cite{Perceive-IR10990319}, and DSwinIR~\cite{DswinIR11304568}, directly learn unified mappings for different degradation combinations. Although they achieve competitive results in several single or relatively simple degradation cases, their performance becomes less stable when multiple degradations coexist. In contrast, agent-based methods consistently achieve higher PSNR and SSIM in most mixed-degradation settings, showing the advantage of degradation-aware tool coordination and adaptive restoration planning.
Among existing agents, MAIR~\cite{MAIRjiang2026multi}, AgenticIR~\cite{agenticir}, HybridAgent~\cite{hybridagentli2026hybrid}, IAMAgent~\cite{wei2026iamagent}, and Restore-R1~\cite{lu2026restorer1} already improve substantially over restoration models. Causal-AgentIR further achieves the best PSNR.
The advantage of Causal-AgentIR is particularly evident in compound degradation scenarios. For H+R+N, it improves PSNR from the best restoration-model result of 30.22 dB to 31.66 dB, yielding a gain of 1.44 dB. These gains indicate that static restoration models struggle to handle interacting degradations, whereas agentic methods can better adapt restoration strategies according to the degradation composition.

Figure~\ref{fig:mixm} further illustrates qualitative comparisons under mixed degradations. Restoration models often fail to simultaneously suppress haze, rain, and noise, leading to residual artifacts, over-smoothing, or color distortion. Existing agents alleviate these issues through tool-based restoration. Causal-AgentIR produces cleaner results with sharper structures and more natural color recovery, benefiting from causal memory-guided planning.

\subsection{Agent Efficiency and Preference Alignment}
\label{subsec:agent_efficiency}

Beyond restoration accuracy, an effective restoration agent should also make reliable decisions with fewer unnecessary corrections and better alignment with user preferences. Therefore, we evaluate different agentic methods using three additional metrics: rollback rate (RR), restoration success rate (SR), and preference alignment rate (PAR).

\textbf{Rollback rate.}
We report the rollback rate (RR), where a lower value indicates fewer ineffective or harmful restoration operations:
\begin{equation}
\mathrm{RR}=\frac{N_{\mathrm{rollback}}}
{N_{\mathrm{execution}}},
\label{eq:rollback_rate}
\end{equation}
where $N_{\mathrm{rollback}}$ is the number of restoration operations explicitly reverted by the agent and $N_{\mathrm{execution}}$ is the total number of executed operations.

\textbf{Restoration success rate.}
We further report the restoration success rate (SR), which measures whether the restored output improves the input without introducing severe artifacts:
\begin{equation}
\mathrm{SR}=\frac{1}{N}
\sum_{i=1}^{N}
\mathbb{I}
\left[
Q(I_i^{\mathrm{out}})
>
Q(I_i^{\mathrm{in}})
\land
A(I_i^{\mathrm{out}})=0
\right],
\label{eq:success_rate}
\end{equation}
where $Q(\cdot)$ denotes the composite restoration quality score, $A(\cdot)$ is the severe-artifact detector, and $\mathbb{I}[\cdot]$ is the indicator function.

\textbf{Preference alignment.}
For preference-conditioned restoration, participants compare anonymized outputs generated under instructions such as ``preserve natural details'', ``prefer stronger artifact removal'', and ``avoid over-enhancement''. The preference alignment rate is defined as:
\begin{equation}
\mathrm{PAR}
=
\frac{N_{\mathrm{agree}}}
{N_{\mathrm{total}}},
\label{eq:preference_alignment}
\end{equation}
where $N_{\mathrm{agree}}$ is the number of comparisons in which the system output agrees with the participant preference. Each comparison is evaluated by at least 50 independent participants.

Table~\ref{tab:agent_efficiency} compares Causal-AgentIR with existing restoration agents. Causal-AgentIR achieves the lowest RR and the highest SR and PAR among the compared methods. The reduced RR indicates that the proposed causal memory helps the planner avoid unreliable restoration operations before execution, thereby reducing reflection and rollback. The improved SR shows that Causal-AgentIR produces successful restoration results more consistently. Moreover, the higher PAR demonstrates that self-evolving memory and human feedback allow the agent to better align restoration behavior with user preferences.

\begin{table}
\centering
\caption{Comparison of agent efficiency and preference alignment.}
\label{tab:agent_efficiency}

\begin{tabular}{cccc}
    \hline
    Method & RR $\downarrow$ & SR $\uparrow$ & PAR $\uparrow$ \\
    \hline
    \hline
    AgenticIR~\cite{agenticir} & 0.32 & 0.84 & \underline{0.88}
    \\
    MAIR~\cite{MAIRjiang2026multi} & 0.36 & \underline{0.90} & 0.80
    \\
    HybridAgent~\cite{hybridagentli2026hybrid} &0.38 &0.88 & 0.82
    \\
    IAMAgent~\cite{wei2026iamagent} & 0.32 & \underline{0.90} & 0.86
    \\
    Restore-R1~\cite{lu2026restorer1} & \underline{0.28} & 0.88 & 0.86
    \\
    \hline
    \rowcolor{gray!20}
    \textbf{Causal-AgentIR} & \textbf{0.12} & \textbf{0.92} & \textbf{0.90} 
    \\
    \hline
\end{tabular}
\end{table}

\subsection{Ablation Studies}

We conduct ablation studies to evaluate the contribution of multi-agent collaboration and self-evolving causal memory. As shown in Table~\ref{tab:abl1}, the baseline without these two components achieves 34.82 dB PSNR, with a rollback rate (RR) of 0.38, a success rate (SR) of 0.82, and a preference alignment ratio (PAR) of 0.76.
Introducing the multi-agent mechanism improves PSNR to 35.08 dB, yielding a gain of 0.26 dB. It also increases SR from 0.82 to 0.88 and PAR from 0.76 to 0.82, indicating that collaborative degradation analysis, planning, tool execution, and critique improve restoration reliability and user-preference alignment. When only self-evolving causal memory is used, the PSNR further increases to 35.22 dB, with a gain of 0.40 dB. Meanwhile, RR is reduced from 0.38 to 0.22, and PAR is improved to 0.86, demonstrating that structured memory retrieval and causal reasoning help avoid ineffective operations and guide more reliable restoration planning.
By jointly incorporating multi-agent collaboration and self-evolving causal memory, Causal-AgentIR achieves the best overall performance, with 35.55 dB PSNR, 0.12 RR, 0.92 SR, and 0.90 PAR. Compared with the baseline, it improves PSNR by 0.73 dB, reduces RR by 0.26, and improves SR and PAR by 0.10 and 0.14, respectively. These results show that the two components are complementary, multi-agent collaboration improves task decomposition and execution, while causal memory provides reusable knowledge for more stable and preference-aligned planning.

\begin{table}
    \centering
    \caption{Ablation study on individual components of Causal-AgentIR.}
    \label{tab:abl1}
    \resizebox{\linewidth}{!}{
    \begin{tabular}{ccccccc}
    \hline
        Method& Multi-Agent & Self-Evolving &RR $\downarrow$ &SR $\uparrow$ &PAR $\uparrow$ &  PSNR $\uparrow$ 
         \\
         \hline
         \hline
         (a) & & &0.38 &0.82 &0.76 &  34.82
         \\
         (b) &\ding{52}& &0.36 &0.88 &0.82& 35.08 
         \\
          (c) & &\ding{52}  &0.22 &0.86 &0.86& 35.22 
         \\
        (d)  &\ding{52} &\ding{52}  &0.12 &0.92 &0.90 & 35.55 
         \\
         \hline
    \end{tabular}}
\end{table}

\begin{figure}
    \centerline{\includegraphics[width=1\linewidth]{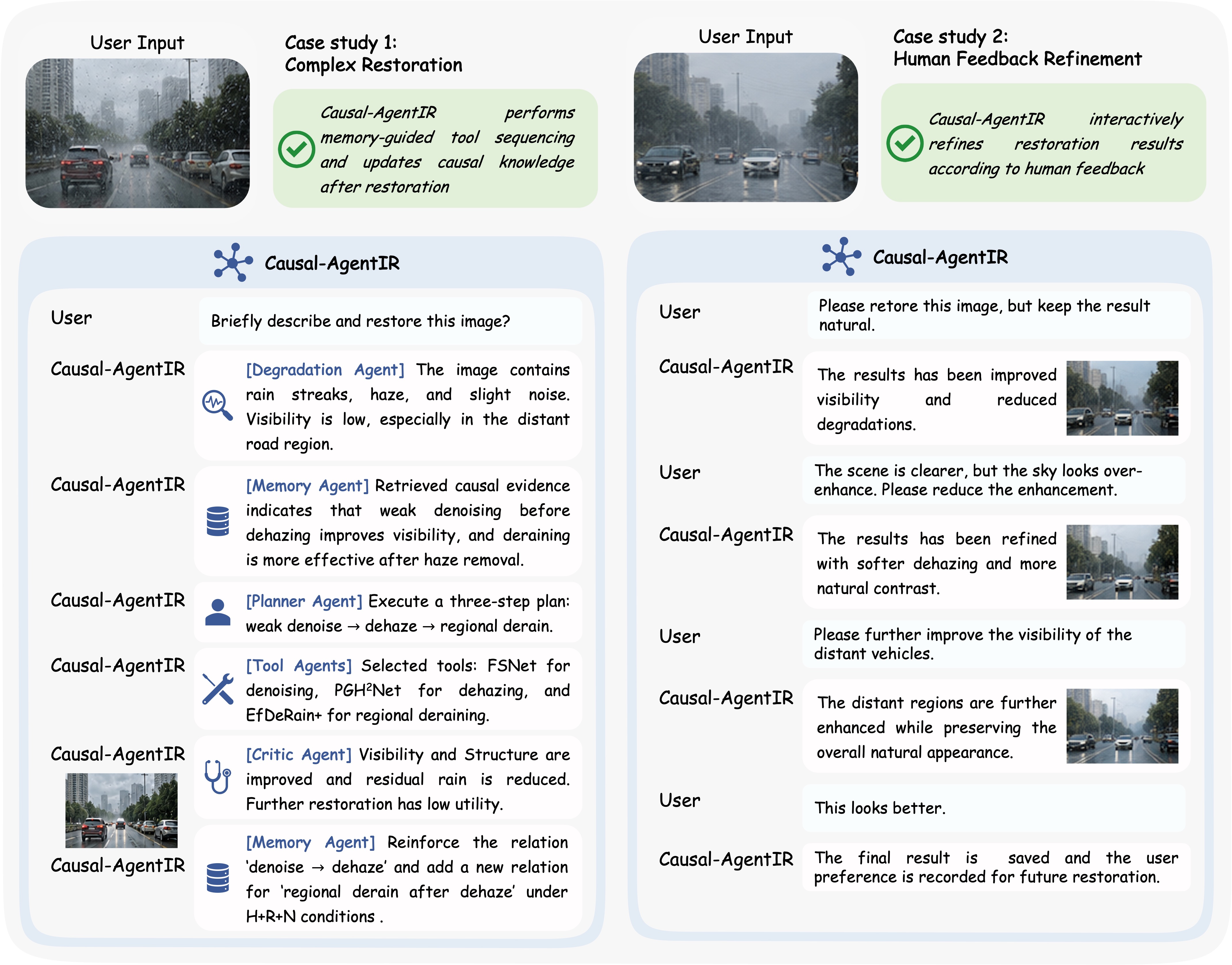}}
\caption{Case studies of Causal-AgentIR. 
}
\label{fig:case_study}
\end{figure}

\subsection{Case Study}
\label{subsec:case_study}

To further illustrate how Causal-AgentIR performs restoration through multi-agent collaboration, causal memory, and human feedback, we provide two representative case studies in Figure~\ref{fig:case_study}. Instead of presenting each internal agent as an isolated module, the restoration process is shown as a unified dialogue of Causal-AgentIR, where different agent roles are explicitly marked in the responses.

In the first case, the input image contains more complex real-world degradations, including rain streaks, haze, and slight noise. The degradation agent first identifies the degradation composition and observes that visibility is low, especially in distant road regions. The memory agent then retrieves causal evidence indicating that weak denoising before dehazing can improve visibility, and that deraining is more effective after haze removal because rain streaks become more distinguishable. Based on this evidence, the planner agent selects a three-step trajectory: weak denoising, dehazing, and regional deraining. The tool agents instantiate this plan by selecting FSNet~\cite{FSNet} for denoising, PGH$^2$Net~\cite{PGH2Netisu2025prior} for dehazing, and EfDeRain+~\cite{efderainguo2025efficientderain+} for regional deraining. After execution, the critic agent confirms that visibility and structure are improved and that residual rain is reduced. Since the expected utility of further restoration becomes low, the process is terminated. The memory agent further reinforces the relation between denoising and subsequent dehazing and adds a new causal relation indicating that regional deraining after dehazing is effective under H+R+N conditions.

The second case demonstrates human feedback refinement. The user first asks Causal-AgentIR to restore a rainy and hazy image while preserving a natural appearance. After the initial restoration, the user observes that the scene becomes clearer but the sky is over-enhanced. Causal-AgentIR then softens the dehazing strength and adjusts the contrast to produce a more natural result. When the user further requests better visibility of distant vehicles, Causal-AgentIR selectively enhances distant regions while preserving the overall appearance. After the user confirms that the result is satisfactory, the final output is saved and the user preference is recorded for future restoration. This case shows that Causal-AgentIR can incorporate human feedback during interaction and align restoration behavior with user-specific preferences.

\begin{figure}
    \centerline{\includegraphics[width=1\linewidth]{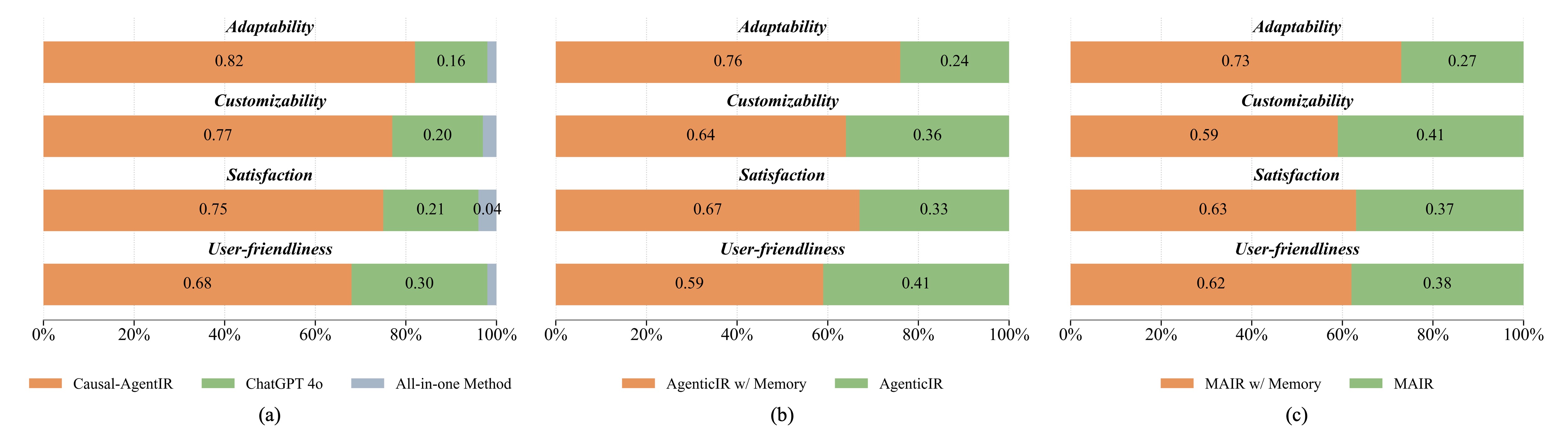}}
	\caption{User study results following the setting of IAMAgent~\cite{wei2026iamagent}. (a) Causal-AgentIR is compared with ChatGPT 4o and an all-in-one restoration method. (b)and (c) Representative restoration agents are compared with their memory-enhanced variants.}
\label{fig:user_study}
\end{figure}

\subsection{User Study}

Following the experimental setting of IAMAgent~\cite{wei2026iamagent}, we conducted a small-scale user study to evaluate the interactive restoration ability of Causal-AgentIR. Specifically, 50 participants from different backgrounds were invited to assess the compared systems using the same four dimensions as IAMAgent, including adaptability, customizability, satisfaction, and user-friendliness. The evaluation followed the same protocol as IAMAgent. Satisfaction was evaluated in a blind manner based on the restored results, while adaptability, customizability, and user-friendliness were evaluated through non-blind interaction, since these dimensions depend on the users' direct experience with system responses and controllability.

Figure~\ref{fig:user_study} reports the user preference results. As shown in Figure~\ref{fig:user_study}(a), Causal-AgentIR is preferred over ChatGPT 4o and the all-in-one method across all four dimensions. It achieves the highest preference ratios in adaptability, customizability, satisfaction, and user-friendliness, indicating that users consider Causal-AgentIR more capable of adapting to different degradation conditions, following user instructions, and producing satisfactory restoration results.
We further evaluate the effect of memory by comparing several representative restoration agents with their  memory-enhanced variants. As shown in Figure~\ref{fig:user_study}(b) and (c), introducing memory consistently improves user preference for AgenticIR and MAIR. The improvement is especially clear in adaptability and satisfaction, suggesting that memory helps restoration agents make more reliable decisions under diverse image conditions. Compared with static or purely interaction-driven agents, memory-enhanced agents can better reuse previous restoration experience and provide more stable responses to user requirements.

\section{Conclusion}
\label{sec:conclusion}

In this paper, we proposed Causal-AgentIR, a hierarchical multi-agent framework with self-evolving causal memory for image restoration. Unlike existing restoration agents that mainly rely on static tool descriptions, predefined priors, or trajectory-level reflection, Causal-AgentIR organizes restoration experience as a structured causal memory graph connecting degradations, regions, tools, actions, quality changes, costs, and user preferences. This design enables graph-based retrieval, multi-hop causal reasoning, and memory-guided planning, allowing the agent to infer effective restoration operations and execution orders under different degradation conditions. A learnable memory evolution mechanism was further introduced to add, update, merge, reinforce, ignore, or discard experience according to observed outcomes and feedback. Extensive experiments demonstrate the effectiveness and generality of Causal-AgentIR.

\bibliographystyle{IEEEtran}
\bibliography{ref}




\vfill

\end{document}